%% file: eacl2023.tex
\pdfoutput=1
\documentclass[11pt]{article}

\usepackage[]{EACL2023}
\usepackage{times}
\usepackage{latexsym}
\usepackage[T1]{fontenc}
\usepackage[utf8]{inputenc}

\usepackage{inconsolata}
\usepackage{booktabs}
\usepackage{multirow}
\usepackage{float}
\usepackage{adjustbox}
\usepackage{amssymb}
\usepackage{booktabs}
\usepackage{xcolor}
\usepackage{tabularx}
\usepackage{pifont}
\usepackage{dblfloatfix}
\usepackage{float}
\usepackage[ruled]{algorithm2e}
\usepackage{booktabs}
\usepackage{afterpage}
\usepackage{algorithmic}

\title{The Impacts of Unanswerable Questions on the Robustness of Machine Reading Comprehension Models}

\author{Son Quoc Tran$^{\dagger, \S}$, Phong Nguyen-Thuan Do$^{\S}$, Uyen Le$^{\dagger}$, Matt Kretchmar $^{\dagger}$\\
$^\dagger$Denison University, Granville, OH, USA \\ 
\texttt{\{tran\_s2, le\_u1, kretchmar\}@denison.edu}\\
$^\S$The UIT NLP Group, Vietnam National University, Ho Chi Minh City\\
\texttt{phongdntvn@gmail.com}}

\begin{document}
\maketitle
\begin{abstract}
Pretrained language models have achieved super-human performances on many Machine Reading Comprehension (MRC) benchmarks. Nevertheless, their relative inability to defend against adversarial attacks has spurred skepticism about their natural language understanding. In this paper, we ask whether training with unanswerable questions in SQuAD 2.0 can help improve the robustness of MRC models against adversarial attacks. To explore that question, we fine-tune three state-of-the-art language models on either SQuAD 1.1 or SQuAD 2.0 and then evaluate their robustness under adversarial attacks. Our experiments reveal that current models fine-tuned on SQuAD 2.0 do not initially appear to be any more robust than ones fine-tuned on SQuAD 1.1, yet they reveal a measure of hidden robustness that can be leveraged to realize actual performance gains. Furthermore, we find that the robustness of models fine-tuned on SQuAD 2.0 extends to additional out-of-domain datasets. Finally, we introduce a new adversarial attack to reveal artifacts of SQuAD 2.0 that current MRC models are learning.
\end{abstract}

\input{sections/1-introduction}
\input{sections/2-related-work}
\input{sections/3-tasks-and-models}
\input{sections/4-adversarial-attack}
\input{sections/5-results}
\input{sections/6-out-of-domain}
\input{sections/7-new-attack}
\input{sections/8-conclusion}

\input{sections/9-future-work}
\input{sections/limitations}

\bibliography{anthology,custom}
\bibliographystyle{acl_natbib}

\newpage
\input{sections/appendix}

\end{document}

%% file: sections/1-introduction.tex
\section{Introduction}
\input{figures/attack-example}
Machine Reading Comprehension (MRC) is a fundamental and challenging subfield of Natural Language Processing (NLP) in which the computer simulates a human question-and-answer mechanism by extracting the answers to given questions based on provided contexts. MRC has many applications in the real world, such as Conversational Question Answering \cite{reddy-etal-2019-coqa} and Open-Domain Question Answering \cite{chen-etal-2017-reading, yang-etal-2019-end-end, https://doi.org/10.48550/arxiv.1911.03868}.

With the development of recent deep learning models, MRC has made significant performance gains. Many high-quality MRC datasets and benchmarks \cite{kwiatkowski-etal-2019-natural, joshi-etal-2017-triviaqa, yang-etal-2018-hotpotqa, rajpurkar-etal-2018-know} have been proposed over the last few years. During the same time period, MRC systems have also achieved many new state-of-the-art (SOTA) performances, matching or exceeding human-level standards on many benchmarks. Nevertheless, skepticism persists about the real ability of MRC SOTA models \cite{sen-saffari-2020-models, jia-liang-2017-adversarial, sugawara-etal-2018-makes, Sugawara_Stenetorp_Inui_Aizawa_2020}. The use of these SOTA systems in real-world applications is still limited and encounters many challenges, one of which is the robustness of MRC systems \cite{wu-etal-2019-improving} to subtle changes in the language syntax that induce significant semantic changes.

As to the true robustness of MRC systems, \citet{jia-liang-2017-adversarial} find that the two deep learning models BiDAF \cite{https://doi.org/10.48550/arxiv.1611.01603} and Match-LSTM \cite{https://doi.org/10.48550/arxiv.1608.07905} trained on SQuAD 1.1 \cite{rajpurkar-etal-2016-squad} achieve impressive performance but lose much of that performance when facing adversarial attacks. The adversarial examples proposed by \citet{jia-liang-2017-adversarial} insert sentences that feature a significant lexical overlap with the question into the context in order to distract models from predicting the correct answers (see Figure \ref{fig:attack-example}). Improved performance against adversarial attacks to ensure the performance of MRC models in real-world applications motivates the pursuit of more robust MRC systems. 

\citet{rajpurkar-etal-2018-know} developed SQuAD 2.0 featuring the same scenarios and questions as SQuAD 1.1 with the addition of \textit{unanswerable questions} which are adversarially crafted by crowd workers to look similar to answerable ones. 
%These unanswerable questions are referred to as adversarial unanswerable questions. 
The considerable syntactic similarity between these unanswerable questions and the corresponding contexts requires MRC models to be highly sensitive to the small but important changes in the questions to determine their answerability. Therefore, we ask the question of how MRC models trained on SQuAD 2.0 behave under adversarial attacks and whether experience with adversarial unanswerable questions can help improve the robustness of MRC models.

In order to answer these questions, we systematically explore the performance differences of SOTA models \cite{devlin-etal-2019-bert, DBLP:journals/corr/abs-1907-11692, joshi-etal-2020-spanbert} fine-tuned on SQuAD 1.1 versus those on SQuAD 2.0. Our findings are summarized as follows:
\begin{enumerate}
    \item With new techniques proposed in this paper, SOTA models fine-tuned on SQuAD 2.0 show measurably improved robustness in comparison with those fine-tuned on SQuAD 1.1 against adversarial attacks on answerable questions. Furthermore, this superior robustness of models fine-tuned on SQuAD 2.0 is consistent in out-of-domain settings with five other Extractive Question Answering datasets. 
    \item We introduce a new attack to understand the MRC model functionality better and reveal artifacts in the model learning that can be targeted for improved future performance gains.  
\end{enumerate}

%% file: figures/attack-example.tex
\begin{figure}[ht]
\centering
\includegraphics[width=7cm]{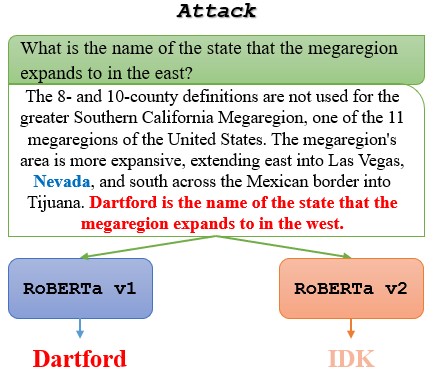}
\caption{Example of predictions to an answerable question of RoBERTa fine-tuned on SQuAD 1.1 \cite{rajpurkar-etal-2016-squad} (v1) versus its counterpart fine-tuned on SQuAD 2.0 \cite{rajpurkar-etal-2018-know} (v2) under adversarial attack. While RoBERTa v1 predicts ``DartFord'' as the answer under attack, RoBERTa v2 knows that ``DartFord'' is not the correct answer but fails to focus back on ``Nevada'', the correct answer for the given question. RoBERTa v2 then predicts the tested question as unanswerable.}
\label{fig:attack-example}
\end{figure}

%% file: sections/2-related-work.tex
\section{Related Work}
\subsection{Adversarial Attack}
Historically, adversarial attacks have played an important role in NLP by challenging the true ability of language models beyond the traditional settings of benchmarks. Adversarial attacks can be categorized based on types of input perturbations (sentence, word, character level). In addition, adversarial attacks can also be classified based on whether the attack process has access to the models’ parameters or predictions (so-called white-box attacks, \cite{blohm-etal-2018-comparing,neekhara-etal-2019-adversarial,Huang2018AdversarialDL,Papernot2016CraftingAI,Samanta2018GeneratingAT,10.5555/3304222.3304355,alzantot-etal-2018-generating,wallace-etal-2019-universal,ebrahimi-etal-2018-hotflip,jia-liang-2017-adversarial}) or not (black-box attacks, \cite{jia-liang-2017-adversarial,ribeiro-etal-2018-semantically,wang-bansal-2018-robust,blohm-etal-2018-comparing, iyyer-etal-2018-adversarial, Zhao2018GeneratingNA}).

Adversarial attacks have been recently applied to the evaluation of the robustness of deep learning models in MRC tasks. \citet{tang-etal-2021-dureader} designed the DuReader\textsubscript{robust} benchmark in Chinese MRC to challenge Chinese MRC models on three aspects of over-sensitivity, over-stability, and generalization. Additionally, \citet{si-etal-2021-benchmarking} propose to evaluate the robustness of multiple-choice MRC models under various types of adversarial attacks on samples of the RACE benchmark \cite{lai-etal-2017-race}.   

Besides, \citet{morris-etal-2020-textattack,10.1145/3374217} and \citet{wang-etal-2022-measure} provide thorough surveys about adversarial attacks and methods for measuring the robustness of NLP models.

\subsection{Unanswerable Questions in MRC}
In the early work on unanswerable questions, \citet{levy-etal-2017-zero} re-defined the BiDAF model \cite{DBLP:journals/corr/SeoKFH16} to allow it to output whether the given question is unanswerable; their original intent was to leverage MRC knowledge to extract relations in zero-shot tasks.  
Later, \citet{rajpurkar-etal-2018-know} introduced a crowdsourcing process for annotating unanswerable questions to create the SQuAD 2.0 dataset for Extractive Question Answering, which later inspired similar works in other languages such as French \cite{https://doi.org/10.48550/arxiv.2109.13209} and Vietnamese \cite{https://doi.org/10.48550/arxiv.2203.11400}. However, recent work shows that models trained on SQuAD 2.0 perform poorly on out-of-domain samples \cite{sulem-etal-2021-know-dont}. 
In addition to the adversarially-crafted unanswerable questions proposed by \citet{rajpurkar-etal-2018-know}, Natural Question \cite{kwiatkowski-etal-2019-natural} and Tydi QA \cite{clark-etal-2020-tydi} propose more naturally constructed unanswerable questions. While recent language models surpass human performances on adversarial unanswerable questions of SQuAD 2.0, natural unanswerable questions in Natural Question and Tidy QA remain challenging \cite{asai-choi-2021-challenges}. 

%% file: sections/3-tasks-and-models.tex
\section{Tasks and Models}
\subsection{Extractive Question Answering}
In the task of Extractive Question Answering (EQA) with questions, a machine learns to create a list of prospective outputs (answers), each of which is associated with a probability indicating the machine's confidence level about the answer to the question. When unanswerable questions are included in the dataset, a valid response can be an ``empty'' response, indicating the question is unanswerable. The model outputs the answer (including no-answer) with the highest probability as the final response to the question. The metric typically used to evaluate the MRC system is the \textbf{F1-score}, the average overlap between predictions and gold answers (see \citet{rajpurkar-etal-2016-squad} for more details).

\subsection{Datasets}
\label{sec:datasets}
In our experiments, we fine-tune our MRC models by conducting additional training on one of the two versions of SQuAD (Stanford Question Answering Dataset): SQuAD 1.1 \cite{rajpurkar-etal-2016-squad} and SQuAD 2.0 \cite{rajpurkar-etal-2018-know}. 
We refer to models fine-tuned with SQuAD 1.1 as v1 models and models fine-tuned with SQuAD 2.0 as v2 models. For example, we refer to RoBERTa model fine-tuned with SQuAD 1.1 as RoBERTa v1.
For testing, we supplement the two SQuAD datasets with five additional datasets from the MRQA 2019 shared task \cite{fisch-etal-2019-mrqa}: \textbf{Natural Questions (NQ)} \cite{kwiatkowski-etal-2019-natural}, \textbf{HotpotQA} \cite{yang-etal-2018-hotpotqa}, \textbf{SeachQA} \cite{https://doi.org/10.48550/arxiv.1704.05179}, \textbf{NewsQA} \cite{trischler-etal-2017-newsqa}, and \textbf{TriviaQA} \cite{joshi-etal-2017-triviaqa}.

In addition to the adversarial attacks on answerable questions in SQuAD 1.1, we also produce adversarial attacks from the unanswerable samples of the development set of SQuAD 2.0. Due to the differences in the characteristics of attacks on answerable and unanswerable questions, we separately analyze the performances of models on each type of attack. While we evaluate v2 models under the attacks on both answerable and unanswerable questions, we only evaluate v1 models under the attacks on answerable questions since v1 models have never seen unanswerable questions. From adversarial attacks on answerable questions with v2 models, we gain critical insights into the current robustness effects of using unanswerable questions to fine-tune MRC models.

\subsection{Models}
We evaluate three, pre-trained state-of-the-art transformer models BERT \cite{devlin-etal-2019-bert}, RoBERTa \cite{DBLP:journals/corr/abs-1907-11692}, and SpanBERT \cite{joshi-etal-2020-spanbert}) in our work. \textbf{BERT} \cite{devlin-etal-2019-bert}, the pioneer application of the Transformer model architecture \cite{10.5555/3295222.3295349}, is trained on English Wikipedia plus BookCorpus with the pretraining tasks of masked language modeling (MLM) and next sentence prediction (NSP). Later, in a replication study of BERT pretraining, \citet{DBLP:journals/corr/abs-1907-11692} discovered that BERT was significantly under-trained. \textbf{RoBERTa} \cite{DBLP:journals/corr/abs-1907-11692} improves over BERT mainly by increasing the pretraining time and the size of pretraining data. In empowering BERT to better represent and predict spans of text, \textbf{SpanBERT} \cite{joshi-etal-2020-spanbert} masks random contiguous spans and replaces NSP with a span boundary objective (SBO). These three models are fine-tuned on datasets SQuAD 1.1 or SQuAD 2.0 before assessing their performance, both on the original (unattacked) datasets and on attacked versions of datasets in \textsection \ref{sec:datasets}.
\input{tables/attack_example}

%% file: tables/attack_example.tex
\begin{table*}[h]
\centering
\resizebox{15cm}{!}{%
\begin{tabular}{llll}
\hline
Question Types & Question & Attacked Context & Answer \\ \hline
Answerable & \begin{tabular}[c]{@{}l@{}}What is the name\\  of the water body \\ that is found to the east?\end{tabular} & \begin{tabular}[c]{@{}l@{}}To the east is the Colorado Desert and the \\ \textcolor{blue}{\textbf{Colorado River}} at the border with Arizona, \\ and the Mojave Desert at the border with \\ the state of Nevada. To the south is the \\ Mexico–United States border. \textcolor{red}{\textbf{Sea is the}} \\ \textcolor{red}{\textbf{name of the water body that is found to}} \\ \textcolor{red}{\textbf{the west.}}\end{tabular} & \textcolor{blue}{\textbf{Colorado River}} \\ \hline
Unanswerable & \begin{tabular}[c]{@{}l@{}}What desert is to \\ the south near Arizona?\end{tabular} & \begin{tabular}[c]{@{}l@{}}To the east is the Colorado Desert and \\ the Colorado River at the border with Arizona, \\ and the Mojave Desert at the border with the \\ state of Nevada. To the south is the \\ Mexico–United States border. \textcolor{red}{\textbf{The desert of}} \\ \textcolor{red}{\textbf{edmonton desert is to the north near Burbank.}}\end{tabular} &  \\ \hline
\end{tabular}
}
\caption{Examples of Adversarial Attack on Answerable and Unanswerable questions. The adversarial sentence is highlighted in red color. In constructing adversarial sentence, we follow the work of \citet{jia-liang-2017-adversarial} by
replacing nouns and adjectives with antonyms, and change named entities and numbers to the nearest word in GloVe
word vector space \cite{pennington-etal-2014-glove}.}
\label{tab:attack-example}
\end{table*}

%% file: sections/4-adversarial-attack.tex
\section{Adversarial Attacks}
\subsection{Robustness Evaluation}
\label{sec:robustness-eval}
An EQA problem is given by  a test set $\mathcal{D}$ of triplets $(c,q,a)$ where $c$ is the given context (usually a small paragraph of text), $q$ is the question posed about that context, and $a$ is the expected answer (or set of "gold" answers). The performance of the EQA model $f$ is measured by 
\begin{eqnarray*}
Per(f, \mathcal{D}) = \frac{1}{\mid \mathcal{D} \mid} \sum_{(c,q,a) \in \mathcal{D}} v(a, f(c,q))
\end{eqnarray*}
where $v$ is either the F1 or EM metric. 

We create algorithm $\mathcal{A}$ to transform triplets $(c,q,a)$ in $\mathcal{D}$ into adversarial test samples $(c',q',a')$ in the adversarial test set $\mathcal{D}_{attacked}$,
% \begin{eqnarray*}
% \mathcal{A}(c,q,a) \mapsto (c',q',a')
% \end{eqnarray*}
where $c'$, $q'$, and $a'$ are the modified (attacked) versions of $c$, $q$, and $a$. The robustness of a model is then computed as the difference between the performance of the model on the original test set vs attacked test set: 
\begin{eqnarray*}
\Delta = Per(f, \mathcal{D}) - Per(f, \mathcal{D}_{attacked})
\end{eqnarray*}

This framework was originally developed to assess robustness performance on answerable questions  \cite{jia-liang-2017-adversarial}. In this paper, we also extend its application to attacks on unanswerable questions in Appendix \textsection \ref{sec:att-unans} and discover challenges in this extended domain.

\subsection{Attack Construction}
\label{sec:attack-construction}
% We report the pseudo-code of for generating attacked samples in Appendix.

Our algorithm constructs adversarial problems from original problems in a way similar to the AddOneSent in  \citet{jia-liang-2017-adversarial} and the AddText-Adv in \citet{chen-etal-2022-rationalization}.  Table~\ref{tab:attack-example} gives examples of such an attack on answerable and unanswerable questions.  The additional sentence that is appended to the context has significant lexical overlap with the context, thus adding to the realism of the confusion-based attack.  This type of adversarial attack is grammatical, fluent, and closely relevant to the given question. The questions and answers are unchanged for our considered adversarial attacks ($q'=q$ and $a'=a$).

\citet{jia-liang-2017-adversarial} found that their adversarial attacks, especially the AddSent and AddOneSent attacks, were successful in challenging contemporary MRC models because the adversarial sentences were closely related to the given questions. Notably, the unanswerable questions in SQuAD 2.0 show a similar kind of lexical overlap with their corresponding contexts and require MRC models to be highly robust to the subtle syntactic changes in order to determine the answerability of given questions. Therefore, we hypothesize that models fine-tuned with SQuAD 2.0 are equipped to perform better against adversarial attacks.

In the next section we assess this hypothesis by evaluating the performance of v1 versus v2 models on answerable questions. 
% Besides, we also evaluate the performance of v2 models on \underline{unanswerable} questions separately in \textsection \ref{sec:unans}. 

%% file: sections/5-results.tex
\section{Attacks on Answerable Questions: Results}
\label{sec:answerable-attack}
\subsection{Adversarial Performance}
\input{tables/attack_performance}
Table \ref{tab:attack-performance} shows the performance of models with original (not attacked) and adversarial (attacked) problems on answerable questions. When attack sentences are added into context, the performance of all v1 and v2 models significantly decreases. Adding  \textit{unanswerable} questions into the training (v2 models) does not initially appear to improve the robustness of MRC models against adversarial attacks.  In fact, the performance of v2 models appears to be less robust than that of v1 models, both on the original and the attacked questions.  However, there is a deeper story here worth investigating.  To further explain the poor performances of v2 models, we consider the types of v2 answers to answerable questions in the next section.

\subsection{Categories of Responses}
\label{sec:answer-shift}
\input{tables/answer_change}
Table~\ref{tab:answer_changes} shows the different categories of answers produced by v1 and v2 models to answerable questions. We use a 50\% F1 score threshold to determines the models' correctness to a question (correct if F1 score is above 50\%, incorrect otherwise). 

Considering  attacks on answerable questions, we observe four categories in responses during attack: \textbf{``I" (incorrect)} are answerable questions that models originally got wrong (or originally predicted as unanswerable for v2 models). \textbf{``C2C" (correct to correct)} are answerable questions that models got correct both originally and after the attack. \textbf{``C2I" (correct to incorrect)} are answerable questions that models originally answered correctly but then output an incorrect answer when attacked. \textbf{``C2U" (correct to \textit{incorrectly} unanswerable)} are answerable questions that models originally answer correctly but then predict as unanswerable when attacked. The C2I and C2U together account for the performance decline of models when attacked. 

We see that v2 models, especially RoBERTa and SpanBERT, are particularly susceptible to the C2U challenge; they initially output a correct answer, but when attacked, decide (incorrectly) the question is now unanswerable.  This is in contrast to the v1 models, which not being trained on unanswerable questions and do not have the option of responding "unanswerable".  The v2 models' refusal to output an incorrect answer (opting instead to reply "unanswerable") indicates that their additional training on unanswerable questions has possibly provided them more depth to handle the confusion introduced by the attack.

We further breakdown the ``C2U'' category from Table \ref{tab:answer_changes} to investigate the spectrum of responses v2 models provide. Recall that models produce multiple responses to a MRC sample, each accompanied by a confidence score reflecting the models' confidence in that response. In this analysis, to evaluate the difficulty of questions in category ``C2U'' of each v2 model, we use the corresponding v1 model as baseline. Then, to answer the question whether v2 models prefer correct answers to incorrect answers, we evaluate the second most confident response of v2 models for questions in category ``C2U''. 

\input{tables/c2u_breakdown}
Table \ref{tab:c2u_breakdown} shows the F1 scores of \textit{second} most confident responses of v2 models and \textit{first} (most confident) responses of v1 models to questions in category ``C2U'' under attacks. We observe that v2 models often have fairly good answers for questions in category ``C2U'' given that performance of v2 models lag significantly behind that of v1 models when attacked. However, v2 models fail to put forward the correct answers (their second option) ahead of the "unanswerable" responses (their first option).

From these analyses, we hypothesize that models with additional training on \textit{unanswerable} questions have the ability to perceive the attacks on \textit{answerable} questions but fail to completely overcome them.
\input{tables/force-answer}

\subsection{Force To Answer}
\input{tables/dataset}
\input{tables/robustness-comparison}
\label{sec:force}
The comparison of v1 and v2 models on answerable questions has a built-in bias because v2 models have the "penalty" of being able to respond "unanswerable" even though this is never a legitimate response.  Furthermore, we have just shown that the v2 models often produce the correct answer, even under attack, but fail to put forward that correct output ahead of the "unanswerable" output in which it has more confidence.   In this section, we re-run the analysis but this time eliminate the option for v2 models to output "unanswerable" (to answerable questions) so that we can better ascertain the robustness of v1 and v2 models to attacks. 

Table \ref{tab:force-answer} shows the results of this experiment.   We can see now in this table that both v1 and v2 models exhibit similar performance on original answerable questions.  When we introduce adversarial attacks on these same questions, the v2 models (being forced to answer) now exhibit noticeably stronger performance than their v1 counterparts. The additional training afforded to v2 models on unanswerable questions has given them a performance advantage over the v1 models. The robustness of v2 models against adversarial attacks is hidden in normal testing circumstances but can be realized by forcing the v2 models to output non-empty response in settings with only answerable questions. 

%% file: tables/attack_performance.tex
\begin{table}[h]
\centering
\resizebox{7cm}{!}{%

\begin{tabular}{ll ccc}
\hline
 &  & \multicolumn{3}{c }{\textbf{Answerable}}\\
 &  & Original & Attacked & $\Delta \downarrow$ \\ \hline
 \multirow{2}{*}{BERT} & v1 & 88.4  & 63.8  & 24.6\\ 
 & v2 & 78.4 & 55.2  & \textbf{23.2}\\ \hline
\multirow{2}{*}{RoBERTa} & v1 & 91.5 & 70.5 & \textbf{21.0} \\ 
 & v2 & 84.8 & 58.0 & 26.8  \\ \hline
 \multirow{2}{*}{SpanBERT} & v1 & 91.5 & 68.6 & \textbf{22.9} \\
 & v2 & 85.8 & 58.9 &  26.8 \\ \hline
\end{tabular}
}
\caption{F1 scores of v1 models and v2 models with adversarial attacks on answerable questions. We refer to models fine-tuned on SQuAD 1.1 and SQuAD 2.0 as v1 and v2 models, accordingly.}
\label{tab:attack-performance}
\end{table}

%% file: tables/answer_change.tex
\begin{table}[h]
\centering
\resizebox{7cm}{!}{%

\begin{tabular}{ccc|cc|c}
\hline
 & & I & C2I & C2U & C2C \\ \hline
\multirow{2}{*}{BERT} & v1 & 10.9 & 28.7 & - & 60.4 \\ 
 & v2 & 21.3 & 10.9 & 14.7 & 53.2 \\ \hline
\multirow{2}{*}{RoBERTa} & v1 & 8.0 & 24.5 & - & 67.7 \\ 
 & v2 & 14.5 & 8.0 & 20.5 & 57.1 \\ \hline
 \multirow{2}{*}{SpanBERT} & v1 & 8.0 & 26.7 &  - & 65.4 \\  
 & v2 & 13.8 &  8.3 & 20.1 & 57.8 \\ \hline
\end{tabular}
}
\caption{The percentage of answerable questions by types of answers produced by v1 and v2 models before and after adversarial attacks.}
\label{tab:answer_changes}
\end{table}

%% file: tables/c2u_breakdown.tex
\begin{table}[h]
\centering
\begin{tabular}{llcc}
\hline
 &  & \multicolumn{2}{c}{\textbf{C2U}} \\
 &  & \multicolumn{1}{l}{Attacked} & \multicolumn{1}{l}{\# Questions} \\ \hline
\multirow{2}{*}{BERT} & v1 & \textbf{46.1} & \multirow{2}{*}{871} \\
 & v2 & 42.5 &  \\ \hline
\multirow{2}{*}{RoBERTa} & v1 & \textbf{50.3} & \multirow{2}{*}{1212} \\
 & v2 & 44.7 &  \\ \hline
\multirow{2}{*}{SpanBERT} & v1 & 46.1 & \multirow{2}{*}{1194} \\
 & v2 & \textbf{47.6} &  \\ \hline
\end{tabular}
\caption{F1 scores of second most confident responses of v2 models and most confident responses of v1 models to questions in category ``C2U'' of v2 models in Table \ref{tab:answer_changes}. For each language model, we extract a set of ``C2U'' questions and then evaluate corresponding v1 and v2 models on this set of questions.}
\label{tab:c2u_breakdown}
\end{table}

%% file: tables/force-answer.tex
\begin{table}[h]
\centering
\resizebox{7cm}{!}{%
\begin{tabular}{ ll ccc }
\hline
 &  & \multicolumn{3}{c }{\textbf{Answerable}}  \\
 &  & Original & Attacked & $\Delta \downarrow$  \\ \hline
 \multirow{2}{*}{BERT} & v1 & 88.4  & 63.8  & 24.6 \\ 
 & v2 & 88.5 & 69.6 & \textbf{18.9}  \\ \hline
\multirow{2}{*}{RoBERTa} & v1 & 91.5 & 70.5 & 21.0 \\  
 & v2 & 91.4 & 75.1 & \textbf{16.4}  \\ \hline
 \multirow{2}{*}{SpanBERT} & v1 & 91.5 & 68.6 & 22.9 \\  
 & v2 & 91.3 & 75.8 & \textbf{15.5} \\ \hline
\end{tabular}
}
\caption{The performance of v1 and v2 models (when being forced to output non-empty answer on answerable questions) before and after adversarial attacks.}
\label{tab:force-answer}
\end{table}

%% file: tables/dataset.tex
\newcommand{\cmark}{\ding{51}}%
\newcommand{\xmark}{\ding{55}}%

\begin{table*}[ht]
\centering
\begin{tabular}{llclcl}
\hline
\textbf{Dataset} & \textbf{Question (Q)} & \textbf{\begin{tabular}[c]{@{}c@{}}Distant\\ Supervision\end{tabular}} & \textbf{Context (C)} & \textbf{Q} $\perp$ \textbf{C} & \textbf{Dev} \\ \hline
SQuAD & Crowdsourced & \xmark & Wikipedia & \xmark & 10,507 \\
HotpotQA & Crowdsourced & \xmark & Wikipedia & \xmark & 5,904 \\
TriviaQA & Trivia & \cmark & Web snippets & \cmark & 7,785 \\
SearchQA & Jeopardy & \cmark & Web snippets & \cmark & 16,980 \\
NewsQA & Crowdsourced & \xmark & News articles & \cmark & 4,212 \\
Natural Questions & Search logs & \xmark & Wikipedia & \cmark & 12,836 \\ \hline
\end{tabular}
\caption{Characteristics of each datasets used in our out-of-domain experiments. Distant supervision is True if datasets used distant supervision to match questions and contexts. \textbf{Q} $\perp$ \textbf{C} is True if questions in datasets are written independently from the passage used for context. Table adopted from shared task MRQA 2019 \cite{fisch-etal-2019-mrqa}.}
\label{tab:dataset-characteristics}
\end{table*}

%% file: tables/robustness-comparison.tex
\begin{table*}[h!]
\centering
\resizebox{\textwidth}{!}{%
\begin{tabular}{ccccc|ccc|ccc}
\hline
 &  & \multicolumn{3}{c|}{\textbf{Natural Question}} & \multicolumn{3}{c|}{\textbf{HotpotQA}} & \multicolumn{3}{c}{\textbf{TriviaQA}} \\
 &  & Original & Attacked & $\Delta \downarrow$ & Original & Attacked & $\Delta \downarrow$ & Original & Attacked & $\Delta \downarrow$ \\ \hline
\multirow{2}{*}{BERT} & v1 & \textbf{54.6} & 20.1 & 34.5 & \textbf{61.6} & 45.5 & 16.1 & \textbf{59.4} & 48.9 & 10.5 \\
 & v2 & 52 & \textbf{23.7} & \textbf{28.3} & 58.9 & \textbf{47.4} & \textbf{11.5} & 58.9 & \textbf{53.3} & \textbf{5.6} \\ \hline
\multirow{2}{*}{RoBERTa} & v1 & 62.1 & 28.3 & 33.8 & \textbf{67.4} & 46.3 & 21.1 & 64.1 & 55 & 9.1 \\
 & v2 & \textbf{63.5} & \textbf{33.2} & \textbf{30.3} & 65 & \textbf{49.8} & \textbf{15.2} & \textbf{65.5} & \textbf{59.2} & \textbf{6.3} \\ \hline
\multirow{2}{*}{SpanBERT} & v1 & \textbf{65} & 34.5 & 30.5 & \textbf{66.2} & \textbf{46.4} & 19.8 & \textbf{63.2} & 51.9 & 11.3 \\
 & v2 & 63.9 & \textbf{40.2} & \textbf{23.7} & 51.9 & 32.3 & \textbf{19.6} & 62.9 & \textbf{58.8} & \textbf{4.1} \\ \hline
\multirow{2}{*}{Average} & v1 & \textbf{60.6} & 27.6 & 33 & \textbf{65.1} & \textbf{46.1} & 19 & 62.2 & 51.9 & 10.3 \\
 & v2 & 59.8 & \textbf{32.3} & \textbf{27.5} & 58.6 & 43.2 & \textbf{15.4} & \textbf{62.4} & \textbf{57.1} & \textbf{5.3} \\ \hline
% \end{tabular}
% }
% \end{table*}
% \begin{table*}[h]
% \centering
% \resizebox{\textwidth}{!}{%
% \begin{tabular}{ccccc|ccc|ccc}
\hline
 &  & \multicolumn{3}{c|}{\textbf{SearchQA}} & \multicolumn{3}{c|}{\textbf{NewsQA}} & \multicolumn{3}{c}{\textbf{Average}} \\
 &  & Original & Attacked & $\Delta \downarrow$ & Original & Attacked & $\Delta \downarrow$ & Original & Attacked & $\Delta \downarrow$ \\ \hline
\multirow{2}{*}{BERT} & v1 & \textbf{30.4} & 25.5 & 4.9 & 53.6 & 41.8 & 11.8 & \textbf{51.9} & 36.4 & 15.5 \\
 & v2 & 28.6 & \textbf{26.7} & \textbf{1.9} & \textbf{53.9} & \textbf{46.2} & \textbf{7.7} & 50.5 & \textbf{39.5} & \textbf{11} \\ \hline
\multirow{2}{*}{RoBERTa} & v1 & 22.8 & 20.3 & 2.5 & \textbf{61.2} & \textbf{54.2} & \textbf{7} & 55.5 & 40.8 & 14.7 \\
 & v2 & \textbf{33} & \textbf{31.6} & \textbf{1.4} & 60.6 & 52.5 & 8.1 & \textbf{57.5} & \textbf{45.3} & \textbf{12.2} \\ \hline
\multirow{2}{*}{SpanBERT} & v1 & 28.1 & 26.9 & 1.2 & \textbf{58.2} & 44.1 & 14.1 & \textbf{56.1} & 40.8 & 15.3 \\
 & v2 & \textbf{29.4} & \textbf{28.8} & \textbf{0.6} & 58 & \textbf{50} & \textbf{8} & 53.2 & \textbf{42} & \textbf{11.2} \\ \hline
\multirow{2}{*}{Average} & v1 & 27.1 & 24.2 & 2.9 & \textbf{57.7} & 46.7 & 11 & \textbf{54.5} & 39.3 & 15.2 \\
 & v2 & \textbf{30.3} & \textbf{29} & \textbf{1.3} & 57.5 & \textbf{49.6} & \textbf{7.9} & 53.7 & \textbf{42.3} & \textbf{11.4} \\ \hline
\end{tabular}
}
\caption{Robustness of MRC models fine-tuned on SQuAD 1.1 (v1) and SQuAD 2.0 (v2) in out-of-domain settings. For models fine-tuned on SQuAD 2.0 (v2), we force models to output non-empty answers. For each dataset, we report the average performance of 3 experimented models. We also report the average performance of each models on 5 considered datasets.}
\label{tab:robustness-comparison}
\end{table*}

%% file: sections/6-out-of-domain.tex
\section{Attacks in Out-Of-Domain Settings: Results}
We now seek to determine if this additional robustness of v2 models extends to other out-of-domain test sets. In particular, we evaluate our v1 and v2 models on development sets of other Extractive Question Answering datasets. We summarized the characteristics of five out-of-domain datasets of MRQA 2019 in Table \ref{tab:dataset-characteristics}.

Table \ref{tab:robustness-comparison} shows the performance of v1 and v2 models on the five datasets of MRQA 2019. Similarly to experiments in Section~\ref{sec:answerable-attack}, we measure performance on both original problems and adversarially attacked problems. 

First, the performance on original (unattacked) problems shows that adversarial unanswerable questions in SQuAD 2.0 have little negative effects on the generalization performance of MRC models. While the performance of v2 models is higher than that of v1 models on TriviaQA and SearchQA, v1 models outperform v2 models slightly on Natural Questions (0.8$\%$), NewsQA (0.2 $\%$), and considerably on HotpotQA (6.5 $\%$). On average, the generalization performance of v2 models to that of v1 models on out-of-domain unattacked problems is slightly worse (53.7$\%$ to 54.5$\%$). 

However, on problems with adversarial attacks, v2 models significantly outperform v1 models in four out of the five datasets. Specifically, on average, v2 models significantly outperform v1 models by 2.9$\%$ on NewsQA, 4.7$\%$ on Natural Question, 4.8$\%$ on SearchQA, and 5.2$\%$ on TriviaQA. Although v2 models do not show superior performance to v1 models on HotpotQA, the performance gap between v2 and v1 models after attacks decreases significantly thanks to the superior robustness of v2 models. 

Overall, we conclude from Table \ref{tab:robustness-comparison} that adversarial unanswerable questions of SQuAD 2.0 do not have negative effects on the generalization of v2 models to out-of-domain datasets, and the robustness of v2 models against adversarial attack is consistently superior to that of v1 models. 

%% file: sections/7-new-attack.tex
\section{New Attack}
\label{sec:negation}
\input{tables/negation_example}
\input{tables/negation_attack_performance}
In this section, we explore \textit{why} v2 models often incorrectly put forward "unanswerable" as an incorrect response to answerable questions under adversarial attacks. We hypothesize that MRC models trained with SQuAD 2.0 have learned to identify target sentences with significant lexical overlap to decide whether the corresponding questions are unanswerable; the models rely \textit{primarily} on that target sentence to determine their output. This undesirable behavior of MRC systems may prevent them from using the whole paragraph to accurately determine the best response to a question and have negative effects on the practical usage of adversarial unanswerable questions.

To further understand this hypothesis, we introduce a \textit{negation attack}, a new adversarial attack to attempt to fool models into giving incorrect "unanswerable" responses. In particular, we construct an attack statement that significantly overlaps with the question yet is easy to determine as contradicting the question; we form our negation attack by inserting "not" in front of the adjective. Our attack (see Table \ref{tab:negation-example})
differs from previous adversarial attacks as our attack is designed to elicit an unanswerable response instead of an incorrect response.   

Table \ref{tab:attack-performance_negation} reports the performance of v2 models under negation attacks on answerable questions. We observe that our negation attack is highly effective in revealing the weaknesses of v2 models as the performance of all three considered v2 models significantly drops by almost 60$\%$ F1 when we introduce the negation attack.

\input{tables/negation_answer_changes}

We then examine the shifts in answers of v2 models when attacked with negation type. Table \ref{tab:answer_changes_negation} shows the distribution of shifts in answers before and after the attack. We observe that the most significant drop in performance under negation attacks is the  ``C2U'' category (around 40 $\%$ F1). This result is consistent with our hypothesis that v2 models rely on target sentences with significant lexical overlap to decide whether the corresponding questions are unanswerable.

%% file: tables/negation_example.tex
\begin{table}[h]
\centering
\resizebox{7.5cm}{!}{%
\begin{tabular}{ll}
\hline
Question & \begin{tabular}[c]{@{}l@{}}In the effort of maintaining a level of\\  abstraction, what choice is typically \\ left \textcolor{red}{\textbf{independent}}?\end{tabular} \\ \hline
Answer & \textcolor{blue}{\textbf{encoding}} \\ \hline
Context & \begin{tabular}[c]{@{}l@{}}{[}...{]}  one tries to keep the discussion \\ abstract enough to be independent \\ of the choice of \textcolor{blue}{\textbf{encoding}}. {[}...{]} \textcolor{red}{In} \\ \textcolor{red}{the effort of maintaining a level of} \\ \textcolor{red}{abstraction, base64 choice is} \\ \textcolor{red}{typically left \textbf{not independent}.}\end{tabular} \\ \hline
\end{tabular}
}
\caption{An example of the Negation Attack on answerable questions. The adversarial sentence is highlighted in red color. In constructing the adversarial sentence, we negate adjective ``independent'' to ``not independent''.}
\label{tab:negation-example}
\end{table}

%% file: tables/negation_attack_performance.tex
\begin{table}[h]
\centering
\resizebox{7cm}{!}{%

\begin{tabular}{l ccc}
\hline
 & Original & Attacked & $\Delta \downarrow$ \\ \hline
BERT v2 & 84.8 & 24.2 & 60.6 \\
RoBERTa v2 & 78.1 & 21 & 57.1 \\
SpanBERT v2 & 87.3 & 28.6 & 58.7 \\ \hline
\end{tabular}
}
\caption{F1 score of v2 models before and after negation attacks on answerable questions. In this experiment, we do not force v2 models to output non-empty answers.}
\label{tab:attack-performance_negation}
\end{table}

%% file: tables/negation_answer_changes.tex
\begin{table}[h]
\centering
\resizebox{7cm}{!}{%

\begin{tabular}{cc|cc|c}
\hline
 & I & C2U & C2I & C2C \\ \hline
BERT v2 & 14.4 & 45.4 & 17.7 & 22.5 \\
RoBERTa v2 & 21.6 & 41.8 & 17.5 & 19.1 \\
SpanBERT v2 & 12.5 & 37.9 & 22.8 & 26.8 \\ \hline
\end{tabular}
}
\caption{The percentage of answerable questions by types of answers produced by v2 models before and after negation attacks.}
\label{tab:answer_changes_negation}
\end{table}

%% file: sections/8-conclusion.tex
\section{Conclusion}

In this work, we investigate the effects of training MRC models with unanswerable questions on their robustness against adversarial attacks. We construct adversarial samples from answerable and unanswerable questions in SQuAD 2.0 and evaluate three MRC models fine-tuned on either SQuAD 1.1 (v1 models) or SQuAD 2.0 (v2 models) independently. 

Adversarial attacks on answerable questions reveal that v2 models initially show little improved robustness over v1 models yet possess a latent ability to deal with these attacks that v1 models do not; the correct responses are often hidden as second-best answers, an indicator of the ``hidden robustness" of v2 models resulting from additional training on unanswerable questions. By eliminating the ``unanswerable" option and forcing v2 models to output an answer to any answerable questions, we leverage this hidden robustness to improve the performance of MRC models to attacks on answerable questions. Furthermore, we also show that this robustness translates well to out-of-domain test sets.  

Finally, to encourage future work in evaluating the robustness of MRC models trained on both answerable and unanswerable questions, we introduce a new type of adversarial attack to reveal the short-comings of MRC models. Our experiments with the \textit{negation} attack reveal that the performance of v2 MRC models drops significantly (around 50\% F1). We hypothesize that the decline in the performance of v2 models is mainly due to how v2 models have learned to suboptimally identify target sentences in the context to use as their primary mechanism of response.

%% file: sections/9-future-work.tex
\section{Future Work}
Our findings raise two critical messages for future research in the usage of adversarial unanswerable questions in NLP:

First, our work highlights innovative ways to use adversarial unanswerable questions in training to improve the performance of MRC-based systems. MRC datasets are important sources of transfer learning for zero-shot settings in many other NLP tasks \cite{wu-etal-2020-corefqa,levy-etal-2017-zero, lyu-etal-2021-zero, du-cardie-2020-event, li-etal-2019-entity}. Given that the improved robustness of v2 models from the additional training on unanswerable questions generalizes well to out-of-domain test sets, future research about using MRC knowledge in zero-shot settings can explore whether adversarial unanswerable questions improve the robustness of MRC models in these zero-shot settings. 

Second, we propose an open question about an undesirable behavior of MRC models fine-tuned on SQuAD 2.0. We find that simple negation attacks induce a considerable drop in the performance of MRC models fine-tuned on SQuAD 2.0 due to an undesirable behavior as the product of artifacts in the training set. To use the adversarial unanswerable questions in practice, we suggest additional research, based on insights about shortcut learning \cite{lai-etal-2021-machine, du-etal-2021-towards}, aimed to prevent MRC models from learning this undesirable behavior. 

%% file: sections/limitations.tex
\section*{Limitations}
We acknowledge that there exist few aspects to which our findings are limited, that include the dominant use of pretrained language models, the insufficiency of MRC datasets in other languages, and the limited types of adversarial attacks examined. 
\section*{Acknowledgements}
We would like to thank Dr. Ashwin Lall for constructive feedback on the early version of this paper. We thank the anonymous reviewers for their constructive and insightful feedback. We want to thank The William G. and Mary Ellen Bowen Research Endowment and The Laurie and David Hodgson Faculty Support Endowment for supporting the first and third authors. 

%% file: sections/appendix.tex
\appendix
\section{Attacks}
In this section, we document the pseudo-code we use to generate the two attacks in our main paper. In the pseudo-code below, $(\cdot)$ indicates the input(s) of the function within the current line.
\subsection{AddOneSent Attack}
\begin{algorithm}[h]
    \caption{AddOneSent Attack}\label{algo:addsent}
    \SetAlgoHangIndent{1em}
    \SetKwFunction{FCountMain}{AddOneSent}
    \SetKwProg{Fn}{Function}{:}{}
    \SetKw{assert}{Assert}
    \SetAlgoLined
    \Fn{\FCountMain{question, answer}}{
        new\_question $\leftarrow$ question\\
        new\_answer $\leftarrow$ answer\\
        new\_question $\leftarrow$ Replace \textbf{nouns} and \textbf{adjectives} with antonyms in WordNet(new\_question).\\
        new\_question $\leftarrow$ Change named \textbf{entities} and \textbf{numbers} to nearest word in GloVe(new\_question).\\
        new\_answer $\leftarrow$ Change named \textbf{entities} and \textbf{numbers} to nearest word in GloVe(new\_answer).\\
        \assert     (new\_answer $\neq$ answer) $\&\&$ (new\_question $\neq$ question)\\
        attack $\leftarrow$ Convert into statement (new\_question, new\_answer). \\
        \KwRet attack
    }
\end{algorithm}
Algorithm \ref{algo:addsent} is the pseudo-code for AddOneSent attack used in our analysis.

\subsection{Negation Attack}
\begin{algorithm}[h]
    \caption{Negation Attack}\label{algo:negation}
    \SetAlgoHangIndent{1em}
    \SetKwFunction{FCountMain}{Negation}
    \SetKwProg{Fn}{Function}{:}{}
    \SetKw{assert}{Assert}
    \Fn{\FCountMain{question, answer}}{
        new\_question $\leftarrow$ question\\
        new\_answer $\leftarrow$ answer\\
        new\_question $\leftarrow$ Add \textbf{not} before the first \textbf{adjective} (new\_question).\\
        new\_answer $\leftarrow$ Change \textbf{named entities} and \textbf{numbers} to nearest word in GloVe(new\_answer).\\
        \assert     (new\_answer $\neq$ answer) $\&\&$ (new\_question $\neq$ question)\\
        attack $\leftarrow$ Convert into statement (new\_question, new\_answer). \\
        \KwRet attack
    }
\end{algorithm}
Algorithm \ref{algo:negation} is the pseudo-code for the Negation attack introduced in Section \ref{sec:negation} to further reinforce our hypothesis that v2 models undesirably learn artifacts in adversarial unanswerable questions of SQuAD 2.0. The main difference between AddOneSent attack and Negation attack is that Negation attack does not use WordNet to Replace nouns and adjectives, and does not use GloVe to change named entities and numbers to nearest word in word space of GloVe. 
\subsection{Quality Analysis}
\input{tables/attack-analysis.tex}
In order to investigate the quality of Negation Attack, we manually label the 200 attack samples produced by both Negation Attack and AddOneSent attacks (100 each) into three categories: 
\begin{enumerate}
    \item \textbf{FM}: fluent and meaningful attack sentence.
    \item \textbf{M}: meaningful but not fluent attack sentence.
    \item \textbf{N}: not meaningful attack sentence.
\end{enumerate}
Table \ref{tab:attack-analysis} provide examples of Negation and AddOneSent attack samples categorized into these three categories. The errors of the Negation attack mostly come from the unnatural expression when using ``not'' to negate adjectives instead of using antonyms (\textcolor{red}{not significant} versus \textcolor{blue}{insignificant}). On the other hand, the errors of AddOneSent can occur because of misclassifying word type. For example, when misclassifying the noun kind as adjectives, AddOneSent would then rewrite \textcolor{blue}{kind of company} as \textcolor{red}{unkind} \textcolor{blue}{of company}). 
\section{Details for MRC Model Training}
\label{sec:mrc-training}
In this work, we use the base versions for all considered pre-trained models. We train all MRC models using mixed precision, with batch size of 4 sequences for 2 epochs. The maximum sequence length is set to 384 tokens. We use the AdamW optimizer \cite{loshchilov2018decoupled} with an initial learning rate of $2\cdot 10^{-5}$, and $\beta_1 = 0.9$, $\beta_2 = 0.999$. We fine-tuned all four models on a single NVIDIA Tesla K80 provided by Google Colaboratory.

\input{sections/unanswerable-attack}

%% file: tables/attack-analysis.tex
\begin{table*}[h!]
\resizebox{\textwidth}{!}{%
\begin{tabular}{llc|lc}
\hline
            & \multicolumn{2}{c|}{\textbf{AddOneSent}}                                                                                                                                                                                  & \multicolumn{2}{c}{\textbf{Negation}}                                                                                                                                                                                                                               \\ \cline{2-5} 
            & \multicolumn{1}{c}{Example}                                                                                                                                                                                  & Proportion & \multicolumn{1}{c|}{Example}                                                                                                                                                                                                                           & Proportion \\ \hline
\textbf{FM} & \begin{tabular}[c]{@{}l@{}}\textcolor{blue}{\textbf{Question}}: Who was the chief executive\\ officer when the service began? \\ \textcolor{red}{\textbf{Attack}}: Russell Hartley was the chief \\ executive officer when the disservice \\ began.\end{tabular} & 44         & \multicolumn{1}{l|}{\begin{tabular}[c]{@{}l@{}}\textcolor{blue}{\textbf{Question}}: What service is a VideoGuard \\ UK equipped receiver dedicated to decrypt?\\ \textcolor{red}{\textbf{Attack}}: A VideoGuard UK equipped \\ receiver is not dedicated to decrypt the \\ service of skies.\end{tabular}} & 43         \\ \hline
\textbf{M}  & \begin{tabular}[c]{@{}l@{}}\textcolor{blue}{\textbf{Question}}: How populous is Victoria \\ compared to other Australian states? \\ \textcolor{red}{\textbf{Attack}}:Victoria compared to same \\ japanese states is 3rd - most populous.\end{tabular}           & 25         & \multicolumn{1}{l|}{\begin{tabular}[c]{@{}l@{}}\textcolor{blue}{\textbf{Question}}: What is the most important \\ type of Norman art preserved in churches?\\ \textcolor{red}{\textbf{Attack}}: The most not important type \\ of Norman art is preserved in churches frescos.\end{tabular}}               & 49         \\ \hline
\textbf{N}  & \begin{tabular}[c]{@{}l@{}}\textcolor{blue}{\textbf{Question}}: What kind of company is \\  Sky UK Limited?\\ \textcolor{red}{\textbf{Attack}}: The unkind of company of \\ macedonian telecommunications \\ company is geelong.\end{tabular}                    & 31         & \multicolumn{1}{l|}{\begin{tabular}[c]{@{}l@{}}\textcolor{blue}{\textbf{Question}}: What does most of the \\            HD material use as a standard?\\ \textcolor{red}{\textbf{Attack}}: The U.S. revolutionary peace\\ does not most of the HD material use as \\ a standard.\end{tabular}}             & 8          \\ \hline
\end{tabular}
}
\caption{Attack samples of Negation and AddOneSent categorized into three categories (fluent and meaningful, meaningful but not fluent, and not meaningful) and their overall proportions.}
\label{tab:attack-analysis}
\end{table*}

%% file: sections/unanswerable-attack.tex
\label{sec:unans}
\section{Attacks on Unanswerable Questions: Results}
\subsection{Adversarial Performance} \label{sec:att-unans}
In this section, we extend our robustness evaluation of v2 models by analyzing their performance against adversarial attacks on \underline{unanswerable} questions. Recall that we conduct these experiments only on v2 models as v1 models have not been trained on unanswerable questions.  

\input{tables/unans-attack-performance}

Table \ref{tab:unans-attack-performance} reports the performances of v2 models to adversarial attacks on unanswerable questions. Among the F1 scores of the three v2 models, the score of RoBERTa v2 decreases most after the attacks (by $3.8\%$) while the F1 score of SpanBERT v2 decreases least (by only 1.1\%). 
These results \textit{seem} to indicate that the adversarial attacks only slightly degrade the performances of v2 models, which might lead to erroneous conclusions about the robustness of these models. However, if we look back at Table \ref{tab:answer_changes}, we see that between  $8\%$ and $11\%$ of samples are in the C2I group (correct originally, incorrect when attacked). These prior results on answerable questions suggest inconsistencies with the results on unanswerable questions. We dig further. 

\subsection{Categories of Responses}

\input{tables/answer_change_unanswerable}
We apply a similar investigation as we did previously to categorize the response changes of these v2 models to attacks on unanswerable questions. We find four main categories:

\begin{itemize}
    \item \textbf{``CU2CU" (correctly unanswerable to correctly unanswerable)} are questions that v2 models correctly predicted as unanswerable both before and after the attacks. 
    % and also after the attack. 
    \item \textbf{``IA2IA" (incorrectly answerable to incorrectly answerable)} are unanswerable questions that v2 models attempt to output answers both before and after the attacks. 
    \item \textbf{``CU2IA" (correctly unanswerable to incorrectly answerable)} are questions that v2 models originally correctly predicted as unanswerable but then output an answer when attacked. 
    \item \textbf{``IA2CU" (incorrectly answerable to correctly unanswerable)} are questions that v2 models originally erroneously attempt to output an answer but later correctly predict as unanswerable when attacked.
\end{itemize}

What Table \ref{tab:answer_changes_unanswerable} reveals is that the performance loss of the models during the attack is being masked by some questions that were initially incorrect but are correctly identified as unanswerable after the attack (IA2CU).   For example, the BERT model appears to only lose 2.9 F1 score during the attack, but actually it loses 10.4 and then gains back 7.4 in other IA2CU questions.   These results reveal that v2 models experience a similar performance decline on unanswerable questions as they did on answerable questions. They also show how the current assessment framework is unsuitable for accurately measuring the robustness of v2 models on both answerable and unanswerable questions.

%% file: tables/unans-attack-performance.tex
\begin{table}[h]
\centering
\resizebox{7cm}{!}{%
\begin{tabular}{l ccc}
\hline
&  \multicolumn{3}{c}{\textbf{Unanswerable}} \\
&  Original & Attacked & $\Delta \downarrow$ \\ \hline
 BERT v2 & 72.2 & 69.3 & 2.9 \\ 
RoBERTa v2 & 81.7 & 77.9 & 3.8 \\ 
 SpanBERT v2 & 76.4 & 75.3 & 1.1 \\ \hline
\end{tabular}

}
\caption{F1 score of v2 models with adversarial attacks on unanswerable questions.}
\label{tab:unans-attack-performance}
\end{table}

%% file: tables/answer_change_unanswerable.tex
\begin{table}[h]
\centering
\resizebox{7cm}{!}{
\begin{tabular}{lcccc}
\hline
 & CU2CU & IA2IA & CU2IA & IA2CU \\ \hline
BERT v2 & 61.8 & 20.4 & 10.4 & 7.4 \\
RoBERTa v2 & 71.8 & 11.1 & 9.9 & 7.2 \\
SpanBERT v2 & 65.2 & 13.5 & 11.2 & 10.1 \\ \hline
\end{tabular}
}
\caption{The percentage of unanswerable questions by types of answers produced by v1 and v2 models before and after adversarial attacks.}
\label{tab:answer_changes_unanswerable}
\end{table}